# Discovering the Symptom Patterns of COVID-19 from Recovered and Deceased Patients Using Apriori Association Rule Mining

Mohammad Dehghani[*]
*School of Electrical and Computer Engineering,*
*University of Tehran,*
*Tehran, Iran*
dehghani.mohammad@ut.ac.ir

Zahra Yazdanparast
*School of Electrical and Computer Engineering,*
*Tarbiat Modares University,*
*Tehran, Iran*
zahra.yazdanparast@modares.ac.ir

**Abstract**

The COVID-19 pandemic has a devastating impact globally, claiming millions of lives and causing significant social and economic disruptions. In order to optimize decision-making and allocate limited resources, it is essential to identify COVID-19 symptoms and determine the severity of each case. Machine learning algorithms offer a potent tool in the medical field, particularly in mining clinical datasets for useful information and guiding scientific decisions. Association rule mining is a machine learning technique for extracting hidden patterns from data. This paper presents an application of association rule mining based Apriori algorithm to discover symptom patterns from COVID-19 patients. The study, using 2875 patient's records, identified the most common signs and symptoms as apnea (72%), cough (64%), fever (59%), weakness (18%), myalgia (14.5%), and sore throat (12%). The proposed method provides clinicians with valuable insight into disease that can assist them in managing and treating it effectively.

## 1. Introduction

The COVID-19 outbreak, which originated in Wuhan, China, in December 2019, has escalated into a global health crisis [1]. The World Health Organization (WHO) declared it a pandemic on January 30, 2020, due to its rapid spread around the world [2]. As of April 12, 2023, the WHO [3] reports over 762 million cases and 6 million deaths worldwide, with variants like Delta and Omicron posing increased infection risks [4].

Significant global efforts have gone into developing potential therapeutics and vaccines [5]. While widespread vaccination that began in early 2021 brought relief, ongoing debates about vaccine reliability continue to influence public acceptance [6]. Although first-generation vaccines have shown promising results in preventing severe disease and death, their efficacy against all Omicron infections remains unclear [7]. The administration of up to four doses in some regions raises questions about long-term effectiveness and safety [8].

The advent of COVID-19 placed enormous pressure on healthcare and medical systems [9]. There were many challenges that healthcare organizations had to address regarding decision-making and planning in order to control and restrain the outbreak [10]. Rapid identification and monitoring of infected patients became crucial in controlling the spread of the disease [11, 12]. The deployment of artificial intelligence has since enabled more accurate and quicker diagnoses [13]. Machine learning, a subset of artificial intelligence, allows systems to learn without explicit programming [14] and has found widespread applications in healthcare [15] and medical informatics [16], including the search for treatments for life-threatening diseases, such as COVID-19 [17]. Association rule mining (ARM) is a machine learning approach that has found utility in healthcare data mining and medical applications including predicting diseases based on symptoms, identifying drug reactions, disease treatments, and detecting fraud [18]. ARM's simplicity and interpretability, expressed as IF-THEN rules, make it particularly suited to disease detection and treatment selection [19]. Consequently, medical services become more affordable, quicker, and efficient [20]. ARM's ability to uncover hidden patterns between symptoms and diseases makes it useful for tackling healthcare challenges, including the early detection of COVID-19 [17].

Since the start of the outbreak, researchers have been striving to control COVID-19. Although widespread vaccination has helped reduce disease and mortality rates, the requirement for multiple doses casts doubts on their long-term effectiveness. As COVID-19 spread for the first time in 2019, no data was available. Therefore, learning from prior experiences and preparing for potential future outbreaks is now vital. The aim of this research is to uncover hidden associations between COVID-19 signs and symptoms using ARM. This study provides physicians with insights to prepare for possible outbreaks. The results obtained from this research may assist medical practitioners in gaining a deeper understanding of COVID-19.

The remainder of the paper is structured as follows: Section 2 reviews previous research in this area. Section 3 outlines the proposed method. Section 4 presents the results. Section 5 provides an analysis and discussion. Finally, Section 6 concludes the study.

## 2. Related work

The recent increase in medical data generation has been a beneficial to machine learning, which has proven effective at analyzing such data [21]. Given the voluminous data, machine learning techniques can deliver predictions nearly as accurate as those made by human experts [22]. The application of machine learning technology has empowered healthcare workers, including doctors, to solve complex problems which can be time-consuming, difficult, and inefficient when tackled manually [23].

Machine learning has been applied in various areas of biomedicine, such as identification of biomarkers, transcriptomics, drug discovery, and prediction of mortality rates. For instance, Mi et al. [24] employed a permutation-based feature importance test for feature selection and constructed predictive models using support vector machines, random forests, and deep neural networks to identify key biomarkers. Xie et al. [25] investigated the potential of plasma metabolites as diagnostic biomarkers for lung cancer, combining metabolomics with machine learning techniques like K-Nearest Neighbor (KNN), Naive Bayes, Adaboost, support vector machines, random forest, and neural networks. Chang et al. [26] proposed a method named RESEPT, which employs an autoencoder and a convolutional neural network to segment tissue architecture based on spatially resolved transcriptomes.

In the domain of drug discovery, Carracedo-Reboredo et al. [27] provided a comprehensive review of machine learning approaches and trends. Popova et al. [28] presented ReLeaSE, a model that combines a supervised learning algorithm and reinforcement learning for new drug development. Jiang et al. [29] developed DECAF, a method that uses deep neural networks and graph theory for predicting ICU mortality. Physiological entities were represented as nodes and interaction relations as edges to construct a graph. The method simulates the propagation of failure risk between physiological functions, enabling early detection of potential risks. In the medical ICU, Nistal-Nuño [30] used Bayesian network, Naive Bayes network, and XGBoost tree ensemble for mortality prediction.

Various machine learning techniques, such as support vector machines, KNN, decision trees, random forests, Naive Bayes, and deep learning, are used in medical applications. ARM is one of the machine learning techniques that can be used in medicine. Some studies used ARM for predicting heart diseases. Khedr et al. [31] present an ARM method to handle distributed medical data sources and used it to predict heart disease. In Inamdar [32] study, support, confidence and lift are used to fine association between different factors that may contribute to heart attacks. A method based on fuzzy logic and ARM was proposed in [33] to find hidden patterns in the medical records of two hospitals in Spain. Dabla et al. [34] present an approach using ARM to find meaningful patterns in pediatric illness. Lakshmi and Vadivu [35] applied weighted ARM to analyze and predict disease comorbidity. ARM was applied by Mohapatra et al. [36] for analyzing Tuberculosis, an infection disease that causes death for many people around the world. Cui et al. [37] proposed a weighted Apriori algorithm to find hidden rules from disease diagnostic data to help clinical decisions.

Furthermore, researchers used ARM to predict and extract knowledge about other diseases. In [38], Apriori and frequent patterns were used to identify factors associated with malignant mesothelioma. The [39] study proposed a method based on weighted ARM to identify important factor in cancer. They aim to classify the masses in mammography images into benign and malignant classes for early detection of disease. In [40], ARM was used to identify risk patterns for type 2 diabetes incidence. The author of [41] present a method to identify differentially expressed genes in lung cancer and ARM was used to identify any association between connected genes. Using ARM, [42] investigated whether there exist consistent patterns of clinical features and differentially expressed genes in type 2 diabetes mellitus, dyslipidemia, and periodontitis diseases. Yamamoto et al. [43] applied ARM to patient symptoms and medications registered in claims data to identify Adverse drug reactions signals.

In case of COVID-19 pandemic, different approaches based on machine leaning and deep learning have been proposed [1, 17]. These approaches were primarily designed to predict and detect diseases and mortality. Jain et al. [44] used ResNet50 deep learning network to classify X-ray images into viral induced pneumonia, bacterial induced pneumonia and normal cases. In next step, they applied ResNet-101 network on viral induced pneumonia class to detect COVID-19. Ismael and Şengür [45] classified COVID-19 and healthy chest X-ray images using ResNet18, ResNet50, ResNet101, VGG16, and VGG19 networks. An analysis of RT-PCR dataset to detect COVID-19 was conducted by [46] using machine learning algorithms including decision tree, support vector machine, K-means clustering, and radial basis function. In order to predict morality, Moulaei et al. [47] first performed synthetic minority over-sampling technique (SMOTE) to handle the class imbalance. Then they used GainRatio Attribute for feature selection and train a model using J48 decision tree, random forest, KNN, Naive Bayes, XGBoost, logistic regression, and multi-layer perceptron.

Some studies used pattern mining to analyze the COVID-19 genome. An analysis of mutations in COVID-19 genome sequences was conducted by Saqib Nawaz et al. [48] by using sequential pattern mining on a corpus of COVID-19 genome sequences in order to reveal common patterns and relationships between nucleotide bases as well as to analyze mutations in the genome sequences. In their next study [49], they analyzed SARS-CoV-2 genome sequences using alignment-free sequence analysis and a variety of distance measures to find (dis)similarities. Sequential pattern mining is used to identify frequent amino acid patterns and their relationships. Finally, an algorithm for the analysis of mutations in genome sequences is proposed.

The aim of [5] was to find COVID-19 symptom patterns using ARM. Based on their results using Wolfram Data Repository [50], fever, cough, pneumonia, sore throat, and breathing problems were the most frequent symptoms. Singh et al. [51] applied ARM to GitHub Data Repository-made COVID-19 patient data to identify symptom patterns. Their study showed fever, cough and malaise/body pain are the most common symptom. Matharaarachchi et al. [52] used twitter data to discover symptom patterns. They extract tweets about COVID symptoms and by using ARM, identified frequent symptoms which were brain fog, fatigue, breathing/lung issues, heart issues, flu symptoms, depression, and general pains.

## 3. Proposed Method

The proposed method is depicted in Fig 1. Initially, data were gathered from various hospitals in Mashhad. We categorized the dataset into two groups: data from all patients and data from deceased individuals. Subsequently, we selected suitable features from the data of all patients and the deceased. Our intention was to illuminate the differences between the signs and symptoms of recovered patients compared to those who passed away. For this, we utilized the Apriori algorithm [53]. The method was executed using the Python programming language, with the Pandas, Numpy, and Sklearn libraries aiding in data preprocessing. Pyfpgrowth and Mlxtend packages in Python were used to model our method.

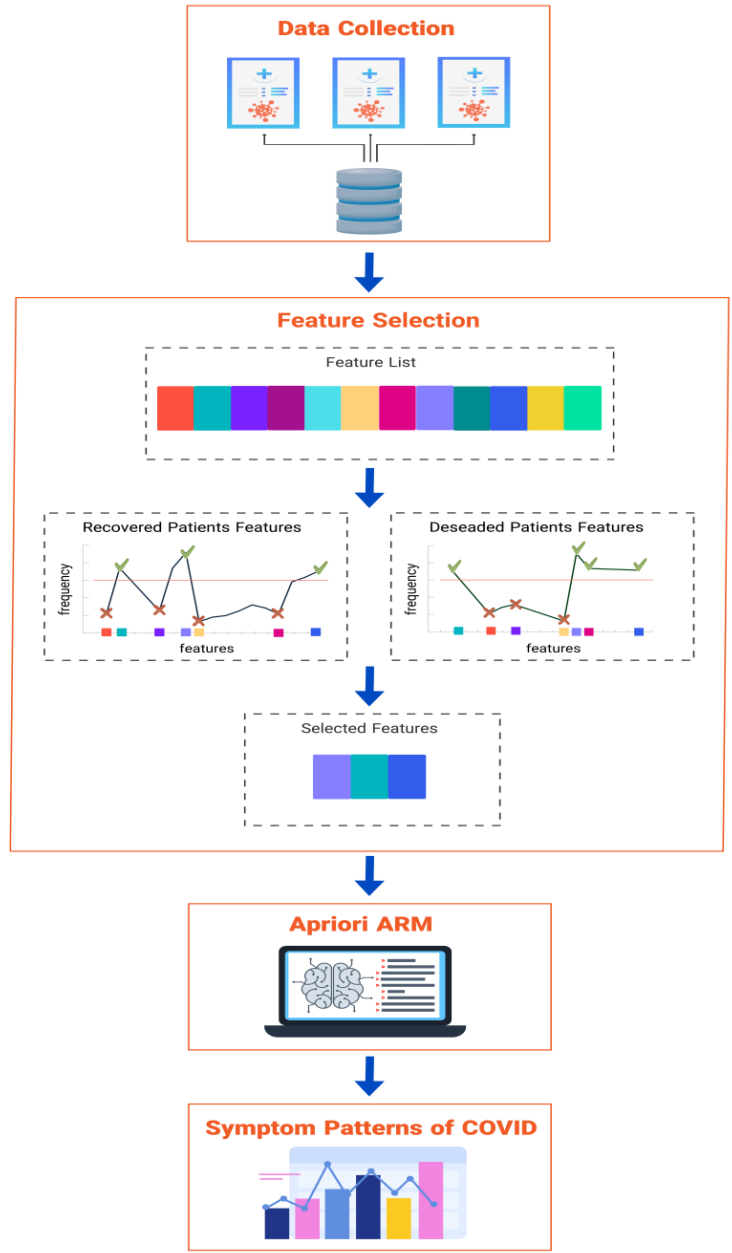

Fig. 1: Our proposed method.

*3.1. Dataset*

We present a comprehensive dataset derived from the medical records of COVID-19 patients treated in three primary hospitals in Mashhad, Iran, from March 2020 through March 2021. This dataset, comprising 2875 individual cases, offers a rich array of clinical characteristics and outcomes associated with COVID-19. The median patient age was 57 years (Standard Deviation ± 21.1 years; Interquartile Range 71 years), with males accounting for 59% of the total population. A strikingly high mortality rate of 24% (693 patients) was observed within this cohort, underscoring the lethal nature of this pandemic.

## 3.2. Feature selection

The dataset includes 34 signs and symptoms, each assigned a value of one (indicating the presence of the symptom) or zero (indicating the absence of the symptom). Not all these symptoms bear equal significance. By employing feature selection, we can consider more relevant features yielding more accurate results. We examined each symptom column, such as 'sore throat', determining the percentage of instances where the symptom was present (indicated by a value of one). This allowed us to gauge the importance of each symptom by its frequency in the entire dataset. After trying different values, we set the threshold at 0.15, an experimental value that appeared suitable. We retained symptoms with a frequency above this threshold and discarded the rest from the dataset. Each symptom and its corresponding frequency are depicted in Fig. 2. For instance, we can overlook 'conjunctivitis' due to its frequency of 0.005. The most frequent signs and symptoms were 'apnea', 'cough', 'fever', 'Ab_Chest_Xray', 'CVD', 'ventilator', and 'weakness'.

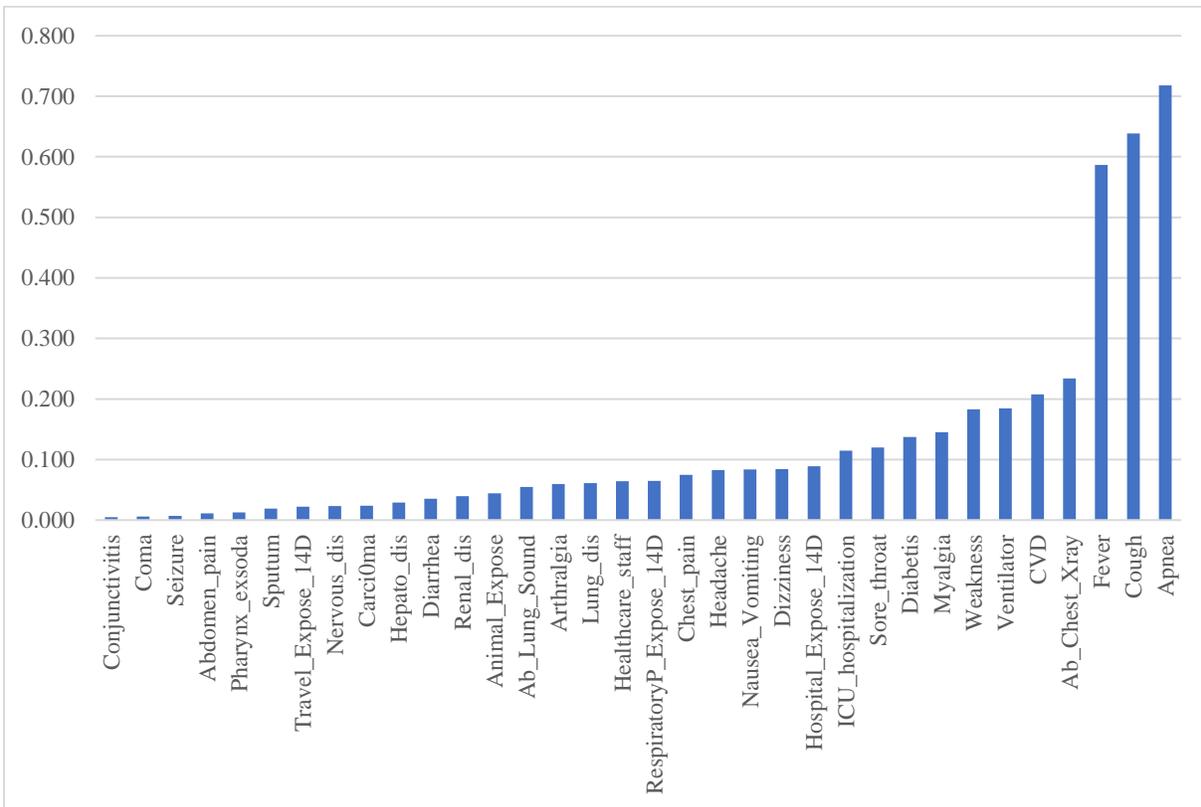

Fig. 2: Symptom frequency of all patients.

We applied the same feature selection method to the data of the deceased patients to ascertain the prevalence of each symptom in this group. We calculated the frequency of signs and symptoms among the deceased and determined the percentage of individuals exhibiting any given symptom, such as 'sore throat'. The optimal threshold for this group was 0.25. Notably, the threshold was determined experimentally in each case and calculated independently. As illustrated in Fig. 3, the most common signs and symptoms among the deceased were 'apnea', 'fever', 'cough', 'ventilator', 'CVD', and 'Ab_Chest_Xray'.

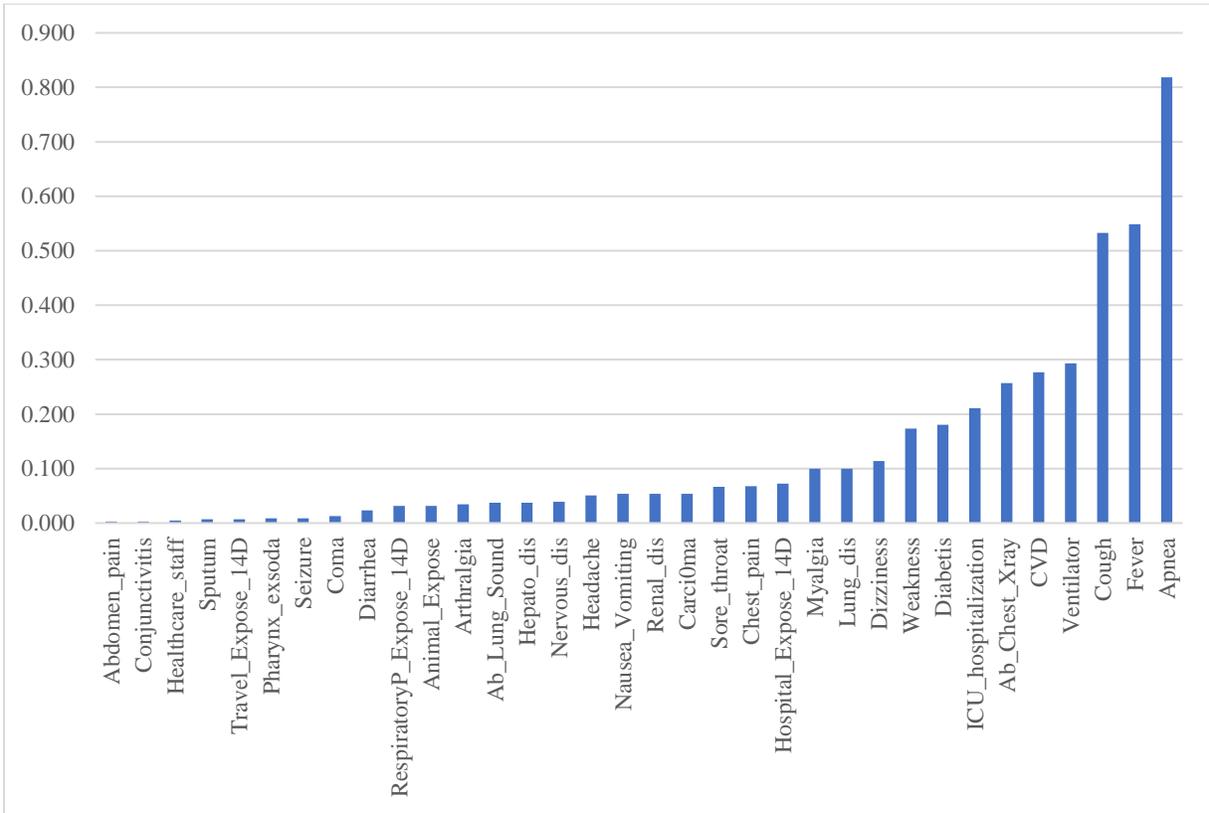

Fig. 3: Symptom frequency of deceased people data.

To improve the accuracy of our results, we combined the two lists of frequent signs and symptoms (from all patients and deceased patients) and eliminated duplicates. The final features were 'apnea', 'cough', 'fever', 'ventilator', 'Ab_Chest_Xray', 'CVD', and 'weakness'. These features had no null values and were used to extract association rules. In Table 1, each sign and symptom is described using Medical Subject Headings (MeSH) [54].

Table 1: Symptom's explanations.

| Symptom | Explanation |
| --- | --- |
| Apnea | A transient absence of spontaneous respiration. |
| Ventilator | A mechanical device that produces or assists pulmonary ventilation. |
| Cough | Air is suddenly expelled from the lungs through a partially closed glottis, that is preceded by inhalation. This protective response allows the trachea, bronchi, or lungs to be cleared of irritants and secretions, or to prevent aspiration of foreign materials into the lungs. |
| Fever | A condition in which the body's temperature is abnormally elevated, usually as a result of a pathological process. |
| Ab_Chest_Xray | Abnormal Chest Xray according to physician diagnosis. |
| CVD | CVD stands for Cardiovascular Disease, which refers to a class of diseases that involve the heart or blood vessels, such as coronary artery disease, heart failure, or stroke. |
| Weakness | A vague complaint of debility, fatigue, or exhaustion attributed to muscle weakness. The weakness can be characterized as subacute or chronic, often |

|  | progressive, and is a manifestation of variety muscle and neuromuscular diseases. |
|---|---|
| Myalgia | Myalgia refers to muscle pain or discomfort. It can be caused by various factors, such as overuse, injury, inflammation, or infection. |
| Diabetes | Diabetes is a chronic disease that occurs either when the pancreas does not produce enough insulin or when the body cannot effectively use the insulin it produces. It results in elevated blood sugar levels, leading to various symptoms and complications. |
| Sore_throat | Sore throat, also known as pharyngitis, is the inflammation of the throat (pharynx). It can be caused by viral or bacterial infections, allergies, or environmental irritants. |
| ICU_hospitalization | ICU hospitalization refers to the admission of a patient to the Intensive Care Unit (ICU) of a hospital. This is usually done when a patient requires close monitoring and specialized care due to a severe or life-threatening condition. |
| Hospital_Expose_14D | Hospital exposure within the last 14 days indicates that an individual has been in a hospital setting or has come into contact with healthcare facilities, potentially increasing the risk of exposure to infectious agents or other health-related factors. |
| Dizziness | Dizziness is a symptom characterized by a sensation of lightheadedness, unsteadiness, or a spinning or swaying motion. It can be caused by various factors, including inner ear problems, medication side effects, low blood pressure, or neurological conditions. |
| Nausea_Vomiting | Nausea and vomiting are common symptoms that can occur due to various causes, such as gastrointestinal infections, motion sickness, food poisoning, pregnancy, or side effects of medications. |
| Headache | Headache refers to pain or discomfort in the head or neck region. It can have various causes, including tension, migraines, sinusitis, or underlying medical conditions. |
| Chest_pain | Chest pain is a symptom that can arise from various causes, ranging from benign musculoskeletal issues to more serious conditions such as heart disease, pulmonary embolism, or gastrointestinal problems. |
| RespiratoryP_Expose_14D | Respiratory exposure within the last 14 days indicates that an individual has been in contact with respiratory pathogens or environments that may increase the risk of respiratory infections or diseases. |
| Healthcare_staff | Healthcare staff refers to individuals working in the healthcare sector who provide medical, nursing, or allied health services to patients in various settings, such as hospitals, clinics, or long-term care facilities. |
| Lung_dis | Lung disease refers to a broad range of conditions that affect the lungs, such as asthma, chronic obstructive pulmonary disease (COPD), pneumonia, or lung cancer. |
| Arthralgia | Arthralgia refers to joint pain or discomfort. It can occur due to various causes, including arthritis, injury, infection, or autoimmune conditions. |
| Ab_Lung_Sound | Abnormal lung sounds, also known as adventitious breath sounds, are atypical sounds heard during the examination of the lungs with a stethoscope. These sounds can indicate underlying respiratory conditions or abnormalities. |
| Animal_Expose | Animal exposure refers to contact with animals, which can include pets, livestock, wildlife, or laboratory animals. It can be relevant when assessing the risk of zoonotic infections or allergies. |
| Renal_dis | Renal disease refers to conditions that affect the kidneys, such as chronic kidney disease, kidney infections, or kidney stones. These conditions can result in impaired kidney function and various symptoms. |
| Diarrhea | Diarrhea is characterized by loose or watery stools, often accompanied by increased frequency and urgency of bowel movements. It can be caused by infections, dietary factors, medications, or underlying gastrointestinal conditions. |
| Hepato_dis | Hepatic disease refers to conditions affecting the liver, such as hepatitis, cirrhosis, or liver cancer. These conditions can impair liver function and lead to symptoms and complications. |
| Carcinoma | Carcinoma refers to a type of cancer that originates from epithelial cells, which are present in the lining of various organs and tissues. Carcinomas can occur in different parts of the body and have different subtypes. |
| Nervous_dis | Nervous system disorders refer to conditions that affect the brain, spinal cord, or peripheral nerves. These disorders can include neurological conditions, such as epilepsy, multiple sclerosis, or Parkinson's disease. |
| Travel_Expose_14D | Travel exposure within the last 14 days indicates that an individual has traveled to regions or countries where there may be different health risks, including infectious diseases or environmental factors. |
| Sputum | Sputum refers to mucus or phlegm that is coughed up from the respiratory tract. It can provide valuable information about respiratory conditions, such as infections, bronchitis, or lung diseases. |
| Pharynx_exsoda | Pharyngeal exudate refers to the presence of pus or other abnormal secretions in the throat or pharynx. It can be a sign of bacterial or viral infections, such as strep throat or tonsillitis. |
| Abdomen_pain | Abdominal pain refers to pain or discomfort in the abdomen, which is the area between the chest and pelvis. It can be caused by various factors, including gastrointestinal issues, inflammation, organ problems, or referred pain from other regions. |
| Seizure | A seizure is a sudden, abnormal electrical activity in the brain that can cause changes in behavior, movements, or sensations. Seizures can be associated with epilepsy or other underlying conditions. |
| Coma | Coma refers to a state of profound unconsciousness where an individual is unresponsive to external stimuli. It is usually caused by severe brain injury, metabolic disturbances, or neurological conditions. |
| Conjunctivitis | Conjunctivitis, also known as pink eye, is the inflammation of the conjunctiva, which is the clear tissue covering the white part of the eye and the inner surface of the eyelids. It can be caused by infections, allergies, or irritants. |

### *3.3. Association Rule Mining (ARM)*

While machine learning classification methods categorize new data and clustering methods identify hidden structures in data [55], ARM discovers patterns, like associations between items in a dataset, to provide useful conclusions and recommendations [56]. ARM was first introduced by Agrawal et al. [57] to extract patterns between products from sales data and discover all the rules that can be used to predict the occurrence of an item based on the occurrence of other items.

In ARM, there are two components: an antecedent and a consequent. The antecedent is an item found within the dataset, while the consequent is an item that appears in conjunction with the antecedent. In a medical context, an association rule between symptoms (or diseases) is expressed as X⇒Y, where X and Y are disjoint sets of symptoms (or diseases), i.e., $X \cap Y = \varphi$ [5]. For example, {Symptom1, Symptom2}⇒{Disease1} in clinical data indicates that patients with Symptom1 and Symptom2 are likely to have Disease1 [58]. An example in cancer disease would be a rule like {Irradiation = No and TumorSize = Large}⇒ {Recurrence = High}, which indicates "if the size of patient tumor is large and have no record of irradiation treatment, then the risk of recurrence of patient cancer is high" [55].

In this study, the rules are formed based on the measures of support, confidence, lift, and leverage which are discussed in the following.

*Definition 1. Support*

Support indicates how often a set of items has been repeated in the dataset [58]. It represents the fraction of records containing X∪Y over the total number of records in the dataset. An association with high support indicates that a significant portion of the dataset applies to the rule (i.e., is frequent) [55].

$$Support\ (X \Rightarrow Y) = \frac{Number\ of\ transaction\ containing\ both\ X\ and\ Y}{Total\ number\ of\ transaction\ in\ dataset} \quad (1)$$

*Definition 2. Confidence*

Confidence is the fraction of transactions containing all items in X and Y from the transactions containing items in X alone [55]. Equation (2) is used to calculate confidence [59].

$$Confidence\ (X \Rightarrow Y) = \frac{Support\ (X \cup Y)}{Support\ (X)} \quad (2)$$

It is necessary for users to specify a minimum level of support and confidence in order to drop infrequent or unconfident rules from the mining process [55].

*Definition 3. Lift*

Lift is used to determine the correlation between X and Y. If lift value is equal to 1, X and Y are independent, the value greater than 1 indicates a positive relationship, the value less than 1 indicates a negative relationship [5]. Equation (3) is used to calculate lift [59].

$$Lift\ (X \Rightarrow Y) = \frac{Support\ (X \cup Y)}{Support\ (X) * Support\ (Y)} \quad (3)$$

*Definition 4. Leverage*

Leverage describes how different is the co-occurrence of X and Y of a rule from independence [60].

$$Leverage\ (X \Rightarrow Y) = Support\ (X \cup Y) - Support\ (X) * Support\ (Y) \quad (4)$$

As a brute-force approach, ARM involves identifying all the possible rules and pruning those that do not satisfy the specified conditions [5]. However, this method is not practical due to its computational complexity, so R. Agrawal proposed a method known as Apriori [53]. The Apriori algorithm uses breadth-first search (BFS) and hash tree to find the most frequent itemsets in a transactional dataset [61]. The advantages of the Apriori algorithm are that it is easy to understand and implement, has a wide variety of derivatives, and it is a popular algorithm in data mining [62].

To extract association rules, we used the Apriori algorithm. We removed patients with only one symptom from the dataset, as these patients are not suitable for creating association rules. We considered each patient as a transaction. In the process of extracting rules, we eliminated some duplicate rules. Symptom transactions are part of the ARM algorithm designed to create frequent itemsets that have at least a threshold set by the user. Therefore, we set a "confidence" threshold of 1, following Nahar et al. [63] approach, since rules are ranked using the "confidence" metric. We also set a minimum support threshold above 0.001 and "lift" greater than 1 for positively correlated rules.

## 4. Results

We analyzed data to investigate the clinical characteristics and outcomes of COVID-19. Apnea was the most prevalent symptom, present in 72% of patients. This finding was also reported by Puram et al. [64], who suggested that central apnea may be a complication in patients with severe COVID-19 infection, and by earlier studies [65, 66]. The next most common signs and symptoms were cough (64%), fever (59%), weakness (18%), myalgia (14.5%), and sore throat (12%). Other symptoms reported in 1-10% of patients included dizziness, nausea, headache, chest pain, arthralgia, diarrhea, sputum production, respiratory distress syndrome, and renal disease. These symptoms were also confirmed by other studies [67-69]. Conversely, symptoms such as seizure and conjunctivitis were reported in less than 1% of patients. As a result of our study, seizures were found to be a rare occurrence, but another study conducted by Keshavarzi et al. [70] found that seizures were among the presenting manifestations of COVID-19 in 0.8% of patients admitted to the hospital for severe illness. Despite our results, conjunctivitis was a common symptom in earlier studies conducted around the world [71-73]. Additionally, 20% of patients had CVD, and 14% had diabetes. This finding indicated a poor prognosis for COVID-19 infection, which was also supported by Gupta et al. [74] in their meta-analysis. Table 2 shows the top rules of dataset extracted by our model. All tables are first sorted based on support and then confidence.

Table 2: Top extracted rules.

| Antecedents | Consequents | Antecedent support | Consequent support | Support | Confidence | Lift | Leverage |
|---|---|---|---|---|---|---|---|
| Fever | Cough | 0.5864 | 0.6386 | 0.4024 | 0.6862 | 1.0746 | 0.0279 |
| Apnea, Cough | Fever | 0.4650 | 0.5864 | 0.2908 | 0.6253 | 1.0662 | 0.0181 |
| Ab_Chest_Xray | Apnea | 0.2337 | 0.7183 | 0.1683 | 0.7202 | 1.0028 | 0.0005 |
| Apnea | Ab_Chest_Xray | 0.7183 | 0.2337 | 0.1683 | 0.2344 | 1.0028 | 0.0005 |
| Ab_Chest_Xray, Apnea | CVD | 0.1683 | 0.2077 | 0.0508 | 0.3017 | 1.4527 | 0.0158 |
| Ab_Chest_Xray, Apnea, CVD | Ventilator | 0.0508 | 0.1843 | 0.0240 | 0.4726 | 2.5636 | 0.0146 |
| Ab_Chest_Xray, Weakness, Fever, CVD | Apnea, Cough | 0.0059 | 0.4650 | 0.0031 | 0.5294 | 1.1384 | 0.0004 |
| Weakness, Apnea, Fever, CVD | Ab_Chest_Xray, Cough | 0.0129 | 0.1301 | 0.0031 | 0.2432 | 1.8699 | 0.0015 |
| Weakness, Cough, Fever, CVD | Ab_Chest_Xray, Apnea | 0.0132 | 0.1683 | 0.0031 | 0.2368 | 1.4069 | 0.0009 |

| | | | | | | |
|---|---|---|---|---|---|---|
| Ab_Chest_Xray, Weakness, Apnea, Fever | Cough, CVD | 0.0160 | 0.1203 | 0.0031 | 0.1957 | 1.6257 | 0.0012 |

Table 2 columns are explained as follow:

- Antecedents: This column represents the symptoms or conditions that are considered as the input or 'if' part of the rule.
- Consequents: This column represents the symptoms or conditions that are predicted or 'then' part of the rule.
- Antecedent Support: It shows the proportion of patients in the dataset that have the specific antecedents.
- Consequent Support: It indicates the proportion of patients in the dataset that have the specific consequents.
- Support: This column represents the proportion of patients in the dataset that exhibit both the antecedents and consequents.
- Confidence: It indicates the likelihood of the consequents occurring given the presence of the antecedents. Confidence is calculated by dividing the support by the antecedent support.
- Lift: Lift measures the strength of the association between the antecedents and consequents. It represents the ratio of observed support to the expected support if the antecedents and consequents were independent.
- Leverage: Leverage quantifies the difference between the observed frequency of the antecedents and consequents occurring together and the frequency that would be expected if they were independent.

These metrics help evaluate the significance and strength of the associations between symptoms or conditions in the dataset. Our three top rules are explained in the following:

1. Rule 1: Fever → Cough
   - Support: 0.4024
   - Confidence: 0.6862
   - Lift: 1.0746
   - Leverage: 0.0279

Rule 1 suggests that when a patient has fever, there is a 68.6% chance they will also experience cough. Pervali et al. [68] reported that patients with lung cancer diagnosed with COVID-19 had cough and fever. However, from our perspective, we are interested in finding out if a symptom has a cause and effect relationship. The lift value greater than 1 indicates a positive association between fever and cough, meaning that the occurrence of fever increases the likelihood of coughing.

2. Rule 2: Apnea, Cough → Fever
   - Support: 0.2908
   - Confidence: 0.6253
   - Lift: 1.0662
   - Leverage: 0.0181

Rule 2 indicates that when a patient experience both apnea and cough, there is a 62.5% chance they will also have a fever. The lift value close to 1 suggests a relatively weaker association between apnea, cough, and fever.

3. Rule 3: Ab_Chest_Xray → Apnea
   - Support: 0.1683
   - Confidence: 0.7202
   - Lift: 1.0028
   - Leverage: 0.0005

Rule number 3 suggests that if a patient has an Ab_Chest_Xray, there is a 72.0% chance they will also have apnea. The lift value close to 1 indicates a weak association between Ab_Chest_Xray and apnea.

The proposed model aims to predict whether a patient with certain symptoms will survive or not. The top rules are listed in Table 3.

Table 3: Top rules of proposed model.

| Antecedents | Consequents | Antecedent support | Consequent support | Support | Confidence | Lift | Leverage |
|---|---|---|---|---|---|---|---|
| Fever, Apnea, Cough | Recovery | 0.291 | 0.759 | 0.225 | 0.774 | 1.02 | 0.004 |
| Ventilator, Apnea, Fever | Death | 0.06 | 0.241 | 0.027 | 0.453 | 1.881 | 0.013 |
| Ab_Chest_Xray, Ventilator, CVD, Apnea | Death | 0.024 | 0.241 | 0.011 | 0.478 | 1.984 | 0.006 |
| Cough, Ab_Chest_Xray, Fever, Ventilator | Death | 0.023 | 0.241 | 0.008 | 0.354 | 1.468 | 0.003 |
| CVD, Ventilator, Weakness | Death | 0.008 | 0.241 | 0.005 | 0.591 | 2.451 | 0.003 |
| Apnea, Ab_Chest_Xray, Fever, Ventilator, CVD, Cough | Recovery | 0.003 | 0.759 | 0.002 | 0.778 | 1.025 | 0 |
| Apnea, Weakness, Ab_Chest_Xray, Ventilator, CVD | Death | 0.003 | 0.241 | 0.002 | 0.667 | 2.766 | 0.001 |
| Apnea, Ab_Chest_Xray, Fever, Ventilator, CVD | Death | 0.006 | 0.241 | 0.002 | 0.389 | 1.613 | 0.001 |
| Apnea, Ab_Chest_Xray, | Death | 0.009 | 0.241 | 0.002 | 0.28 | 1.162 | 0 |

| Antecedents | Consequents | Antecedent support | Consequent support | Support | Confidence | Lift | Leverage |
|---|---|---|---|---|---|---|---|
| Ventilator, CVD, Cough | | | | | | | |
| Apnea, Weakness, Ventilator, CVD, Cough | Death | 0.002 | 0.241 | 0.001 | 0.429 | 1.778 | 0 |

The extracted rules in Table 3 show:

1. Combination of symptoms associated with recovery:
   - Patients who recovered from the disease are likely to have specific symptoms, such as fever, apnea, and cough.
2. Combination of symptoms associated with death:
   - Patients who unfortunately died from the disease are more likely to have certain combinations of symptoms, such as apnea, weakness, Ab_Chest_Xray, ventilator, and CVD.

We also divided the dataset into four groups based on the patient's age: under twenty years, twenty to forty, forty to sixty, and sixty and older. The top rules are presented in Table 4.

Table 4: Top rules based on patient's age.

| Antecedents | Consequents | Antecedent support | Consequent support | Support | Confidence | Lift | Leverage |
|---|---|---|---|---|---|---|---|
| Weakness, <20, Female | Lab_Res_Neg | 0.003 | 0.622 | 0.003 | 1 | 1.609 | 0.001 |
| Ab_Chest_Xray, Cough, <20, Apnea | Lab_Res_Neg | 0.002 | 0.622 | 0.002 | 1 | 1.609 | 0.001 |
| Weakness, Apnea, <20, Female | Lab_Res_Neg | 0.002 | 0.622 | 0.002 | 1 | 1.609 | 0.001 |
| Ventilator, Cough, Female, <20 | Lab_Res_Neg | 0.002 | 0.622 | 0.002 | 1 | 1.609 | 0.001 |
| Weakness, Cough, <20, Female | Lab_Res_Neg | 0.002 | 0.622 | 0.002 | 1 | 1.609 | 0.001 |
| Cough, <20, Fever, Ab_Chest_Xray, Apnea | Lab_Res_Neg | 0.002 | 0.622 | 0.002 | 1 | 1.609 | 0.001 |
| Male, Cough, <20, Ab_Chest_Xray, Apnea | Lab_Res_Neg | 0.002 | 0.622 | 0.002 | 1 | 1.609 | 0.001 |
| 20-40, Ab_Chest_Xray, Weakness, Female | Cough | 0.002 | 0.639 | 0.002 | 1 | 1.566 | 0.001 |
| Male, Cough, <20, Lab_Res_Neg, Ab_Chest_Xray | Apnea | 0.002 | 0.718 | 0.002 | 1 | 1.392 | 0 |
| 20-40, CVD, Ventilator | Apnea | 0.002 | 0.718 | 0.002 | 1 | 1.392 | 0.001 |

| Antecedents | Consequents | Antecedent support | Consequent support | Support | Confidence | Lift | Leverage |
|---|---|---|---|---|---|---|---|
| Ventilator, Ab_Chest_Xray, CVD, 20-40 | Apnea | 0.002 | 0.718 | 0.002 | 1 | 1.392 | 0 |
| Ventilator, CVD, Female, 20-40 | Ab_Chest_Xray | 0.001 | 0.234 | 0.001 | 1 | 4.278 | 0.001 |
| Ventilator, Male, Cough, CVD, Lab_Res_Neg, Fever, >60 | Ab_Chest_Xray | 0.001 | 0.234 | 0.001 | 1 | 4.278 | 0.001 |
| 20-40, Weakness, Female, Lab_Res_Pos, Ab_Chest_Xray | Cough, Fever | 0.001 | 0.402 | 0.001 | 1 | 2.485 | 0.001 |

The findings from Table 4 reveal strong associations between specific combinations of symptoms and outcomes in certain demographic groups. For example, female patients under the age of 20 who have symptoms such as weakness, cough, and Ab_Chest_Xray have a negative lab result. These associations have a confidence of 100%, suggesting a strong correlation. The presence of CVD and ventilator use in the 20-40 age group is strongly correlated with the occurrence of apnea, with a confidence of 100%. Similarly, female patients aged from 20 to 40 who have weakness, cough, and Ab_Chest_Xray are also highly associated with each other. These associations have a confidence of 100% and a lift greater than 1, suggesting a positive relationship. Male patients over 60 who use ventilator, have weakness, cough, CVD, fever, and Ab_Chest_Xray are strongly associated with a negative lab result, with a confidence of 100%. These associations have a confidence of 100% and a lift greater than 1, suggesting a positive relationship.

Table 5: Antecedents associated with the 'recovery' or 'death' consequents.

| Antecedents | Consequents | Antecedent support | Consequent support | Support | Confidence | Lift | Leverage |
|---|---|---|---|---|---|---|---|
| Male, Lab_Res_Neg, <20 | Recovery | 0.0271 | 0.759 | 0.0271 | 1 | 1.3176 | 0.0065 |
| Female, 20-40, Lab_Res_Neg, Fever, Cough | Recovery | 0.0237 | 0.759 | 0.0237 | 1 | 1.3176 | 0.0057 |
| Apnea, 20-40, Male, Lab_Res_Neg, Fever, Cough | Recovery | 0.0223 | 0.759 | 0.0223 | 1 | 1.3176 | 0.0054 |
| Female, Apnea, 20-40, Lab_Res_Neg, Fever, Cough | Recovery | 0.0177 | 0.759 | 0.0177 | 1 | 1.3176 | 0.0043 |
| Male, Fever, <20 | Recovery | 0.0163 | 0.759 | 0.0163 | 1 | 1.3176 | 0.0039 |
| Male, <20, Cough | Recovery | 0.0153 | 0.759 | 0.0153 | 1 | 1.3176 | 0.0037 |
| Male, Lab_Res_Neg, <20, Cough | Recovery | 0.0129 | 0.759 | 0.0129 | 1 | 1.3176 | 0.0031 |

| Antecedents | Consequents | Antecedent support | Consequent support | Support | Confidence | Lift | Leverage |
|---|---|---|---|---|---|---|---|
| Female, Lab_Res_Neg, <20, Cough | Recovery | 0.009 | 0.759 | 0.009 | 1 | 1.3176 | 0.0022 |
| Male, <20, Fever, Apnea | Recovery | 0.0087 | 0.759 | 0.0087 | 1 | 1.3176 | 0.0021 |
| Male, <20, Apnea, Ab_Chest_Xray | Recovery | 0.0045 | 0.759 | 0.0045 | 1 | 1.3176 | 0.0011 |
| Female, <20, Ab_Chest_Xray | Recovery | 0.0042 | 0.759 | 0.0042 | 1 | 1.3176 | 0.001 |
| Female, <20, Ventilator | Recovery | 0.0035 | 0.759 | 0.0035 | 1 | 1.3176 | 0.0008 |
| Cough, Male, <20, Ab_Chest_Xray | Recovery | 0.0024 | 0.759 | 0.0024 | 1 | 1.3176 | 0.0006 |
| Female, <20, Apnea, Ab_Chest_Xray | Recovery | 0.0017 | 0.759 | 0.0017 | 1 | 1.3176 | 0.0004 |
| CVD, Ab_Chest_Xray, Male, 20-40 | Recovery | 0.0014 | 0.759 | 0.0014 | 1 | 1.3176 | 0.0003 |
| Female, Apnea, Lab_Res_Neg, Ventilator, 40-60, Weakness | Death | 0.001 | 0.241 | 0.001 | 1 | 4.1486 | 0.0008 |

Table 5 focuses on the antecedents that are associated with the 'recovery' or 'death' consequents. For example, the combination of factors such as gender, age group, lab result, symptoms (fever, cough), and medical conditions (CVD, apnea) is strongly associated with the 'recovery' outcome, with a confidence of 100% and a lift greater than 1. On the other hand, the combination of factors such as female gender, apnea, negative lab result, ventilator use, weakness, and age group (40-60) is highly associated with the 'death' outcome, with a confidence of 100% and a very high lift. These findings suggest potential predictive relationships between the specified antecedents and the outcomes, but they need further investigation and validation.

Another important investigation aims to identify the presence of apnea or the use of a ventilator among the patients who passed away. This investigation is crucial because some patients may have needed ventilator support, which might not have been available to them. Table 6 shows the results.

Table 6: Apnea or ventilator use among deceased peoples.

| Antecedents | Consequents | Antecedent support | Consequent support | Support | Confidence | Lift | Leverage |
|---|---|---|---|---|---|---|---|
| 20-40, Lab_Res_Pos | Apnea | 0.027 | 0.818 | 0.027 | 1 | 1.222 | 0.005 |
| Female, 40-60, Ab_Chest_Xray | Apnea | 0.014 | 0.818 | 0.014 | 1 | 1.222 | 0.003 |
| Apnea, 20-40, Ab_Chest_Xray | Ventilator | 0.01 | 0.293 | 0.01 | 1 | 3.414 | 0.007 |
| Female, Fever, 20-40 | Ventilator | 0.01 | 0.293 | 0.01 | 1 | 3.414 | 0.007 |

| Female, 20-40, Lab_Res_Pos | Apnea | 0.01 | 0.818 | 0.01 | 1 | 1.222 | 0.002 |
| 20-40, Lab_Res_Pos, Ab_Chest_Xray | Apnea | 0.007 | 0.818 | 0.007 | 1 | 1.222 | 0.001 |
| Weakness, Cough, 20-40 | Apnea | 0.006 | 0.818 | 0.006 | 1 | 1.222 | 0.001 |
| Cough, 20-40, Ab_Chest_Xray | Apnea | 0.004 | 0.818 | 0.004 | 1 | 1.222 | 0.001 |
| Fever, 20-40, Ab_Chest_Xray | Apnea | 0.004 | 0.818 | 0.004 | 1 | 1.222 | 0.001 |
| CVD, Cough, 40-60, Ab_Chest_Xray | Apnea | 0.004 | 0.818 | 0.004 | 1 | 1.222 | 0.001 |

Our scientific investigation reveals strong associations between various factors and the occurrence of apnea, a critical medical condition. Specifically, individuals in the age group of 20-40, those with positive lab results, and females experiencing symptoms such as fever show a strong relationship with apnea, as evidenced by a confidence level of 100% and a lift greater than 1. Moreover, the presence of apnea along with additional factors like Ab_Chest_Xray or fever shows a significant association with the use of ventilators, also supported by a confidence level of 100% and a remarkably high lift.

## 5. Discussion

We applied the ARM method to examine the differences in COVID-19 symptom patterns based on age, gender, and mortality status. We also investigated the prevalence of apnea and ventilator use in the 'death' group. Our study confirmed a relatively higher proportion of apnea, cough, weakness, and sore throat in COVID-19 patients, as reported in a recent systematic review of 148 studies [75]. In this study, we analyzed the dataset to explore the associations related to apnea occurrence and ventilator use among patients. Our findings revealed several important results that enhance our understanding of these phenomena. The symptom pattern of fever, apnea, and cough seems to be indicative of patients who are more likely to recover from the disease, which is consistent with other studies [69]. This suggests that these specific symptoms may be positive prognostic indicators for favorable outcomes. On the other hand, some symptom patterns such as apnea, weakness, Ab_Chest_Xray, ventilator use, and CVD are more frequent in patients who unfortunately died from the disease. A study by Giri et al. [76] shows that compared with non-severe patients, symptoms such as fever, cough, dyspnea, existing comorbidities, and complications are prevalent in severe COVID-19 patients. These findings highlight the importance of these symptoms as potential negative prognostic indicators. Using Medline/PubMed, Scopus, and Google Scholar articles from January 1, 2020, to April 2, 2020, the [77] study found that the most common symptoms in COVID-19 patients are: fever, cough, fatigue, dyspnea, and sputum.

Demographic factors also influence the associations between symptom patterns and outcomes. This finding is supported by other studies in this field that show males are more likely than females to be infected with COVID-19, and that the most common signs and symptoms are fever, breathing difficulty, malaise, dry cough and chest pain [78, 79]. Moreover, our analysis shows that the presence of CVD and ventilator use in the age group of 20-40 is strongly correlated with apnea occurrence [80]. Similarly, the combination of weakness, cough, and Ab_Chest_Xray in females aged 20-40 is highly associated with each other. Although these findings are not reported in any other studies, they may help to identify rare combinations of disease and possible outcomes. The associations between symptom patterns and outcomes are supported by a confidence level of

100%, indicating a high degree of certainty. Additionally, the lift values greater than 1 further emphasize the positive relationships between these factors and their respective outcomes.

These findings provide substantial information about the interplay between patient characteristics, symptoms, and apnea occurrence, as well as the subsequent need for ventilator use. These results can facilitate early identification, risk assessment, and appropriate management of patients at risk of apnea or those requiring ventilator support.

## 6. Conclusion

In this paper, we applied ARM techniques to identify symptom patterns for COVID-19. Features are selected based on recovered and deceased patients. The most frequent symptoms in our study included apnea, cough, fever, weakness, myalgia, and sore throat. Apnea, cough, and fever are present in approximately 60% of patients. Hence, if a patient exhibits these symptoms, then they are likely to have COVID-19. According to the results of a variety of articles, fever and cough are two of the most commonly reported symptoms.

Our study revealed significant associations between various factors and apnea occurrence, as well as ventilator use. These findings add to the existing knowledge base and have implications for clinical decision-making and patient management strategies.

We acknowledge the limitations of our study. The analysis was based on the provided dataset, which may not fully capture the diversity of patient populations. Moreover, the generalizability of our findings to other settings should be further validated through larger-scale studies. Future research with larger and more diverse populations is needed to confirm our findings and investigate underlying mechanisms in more detail.

Future work can be conducted based on other type of ARM algorithms. In this research we only applied Apriori algorithm, other algorithms such as Frequent Pattern Growth [81] and equivalence class clustering and bottom-up lattice transversal algorithm (Eclat) [82] can be used for pattern discovery. These two algorithms have the main disadvantage of consuming enormous memory if the data set contains many transactions [5].

This study determined the threshold for feature selection based on experimental results. Future experiments can utilize greedy search to determine the appropriate threshold. It is also possible to calculate a correlation between signs and symptoms and select features based on this information.

Although COVID-19 may have been controlled by widespread vaccination, it was an unexpected disease that caused chaos for a long time. Recent studies have questioned the effectiveness of vaccines, especially against variants of COVID-19. This situation requires further research on COVID-19 data and the use of experience to prepare for possible new variants of COVID-19 or similar diseases in the future.